\documentclass{ecai} 

\usepackage{latexsym}
\usepackage{amssymb}
\usepackage{amsmath}
\usepackage{amsthm}
\usepackage{booktabs}
\usepackage{enumitem}
\usepackage{graphicx}
\usepackage{color}
\usepackage{multirow}
\usepackage{natbib}
\usepackage{comment}

\newcommand{\BibTeX}{B\kern-.05em{\sc i\kern-.025em b}\kern-.08em\TeX}

\begin{document}


\begin{frontmatter}




\title{GMM-Based Comprehensive Feature Extraction and Relative Distance Preservation For Few-Shot Cross-Modal Retrieval}


\author[1]{\fnms{Chengsong}~\snm{Sun}\thanks{Corresponding Author. Email: scs2019@pku.edu.cn}}
\author[2]{\fnms{Weiping}~\snm{Li}}
\author[3]{\fnms{Xiang}~\snm{Li}} 
\author[4]{\fnms{Yuankun}~\snm{Liu}} 
\author[5]{\fnms{Lianlei}~\snm{Shan}}

\address[1,2,3,4]{School of Software and Microelectronics, Peking University}
\address[5]{University of Chinese Academy of Sciences}


\begin{abstract}
Few-shot cross-modal retrieval focuses on learning cross-modal representations with limited training samples, enabling the model to handle unseen classes during inference. Unlike traditional cross-modal retrieval tasks, which assume that both training and testing data share the same class distribution, few-shot retrieval involves data with sparse representations across modalities. Existing methods often fail to adequately model the multi-peak distribution of few-shot cross-modal data, resulting in two main biases in the latent semantic space: intra-modal bias, where sparse samples fail to capture intra-class diversity, and inter-modal bias, where misalignments between image and text distributions exacerbate the semantic gap. These biases hinder retrieval accuracy. To address these issues, we propose a novel method, GCRDP, for few-shot cross-modal retrieval. This approach effectively captures the complex multi-peak distribution of data using a Gaussian Mixture Model (GMM) and incorporates a multi-positive sample contrastive learning mechanism for comprehensive feature modeling. Additionally, we introduce a new strategy for cross-modal semantic alignment, which constrains the relative distances between image and text feature distributions, thereby improving the accuracy of cross-modal representations. We validate our approach through extensive experiments on four benchmark datasets, demonstrating superior performance over six state-of-the-art methods.
\end{abstract}

\end{frontmatter}


\section{Introduction}
The swift expansion of multimedia data and continuous progress in internet technologies have highlighted the crucial necessity for cross-modal retrieval. Despite the notable advancements made by traditional deep-learning-based solutions, they are highly reliant on large-scale annotated datasets, which pose two main challenges for practical application: first, the rapidly increasing volume of data drives up annotation costs; second, conventional models have difficulty with few-shot generalization, especially when dealing with long-tailed data distributions \cite{zhou2023state}. To address these problems, few-shot cross-modal retrieval has emerged as a vital research area, focusing on enabling models to adapt to classes not seen during training \cite{wang2025cross}.

Current research on cross-modal retrieval can be categorized into four main technical approaches. The first approach directly employs general cross-modal retrieval methods, such as Canonical Correlation Analysis (CCA) \cite{hotelling1992relations} and Stacked Cross Attention Networks (SCAN) \cite{lee2018stacked}. These methods are based on the assumption of single-peak distribution across modalities and rely on densely annotated data to optimize alignment models. However, in few-shot scenarios, this single-peak data distribution assumption contradicts the multi-peak distribution nature of real-world data, which may lead to feature space collapse and a decline in performance. The second approach focuses on the transfer of pre-trained knowledge, with models like Pro-CLIP \cite{cao2024pro} and Eva-CLIP \cite{sun2023eva} serving as examples. These models utilize prompt fine-tuning or knowledge distillation to transfer semantic priors from large-scale models, such as CLIP \cite{radford2021learning}, to few-shot tasks. Although they reduce the dependence on annotations, their performance is restricted by the long-tail bias in pre-trained data, impeding their capacity to capture fine-grained features for out-of-distribution categories. The third approach harnesses generative data augmentation techniques, such as CM-GAN \cite{kang2023cm} and MDVAE \cite{tian2022multimodal}, which use Generative Adversarial Networks (GAN) and Variational Autoencoders (VAE) to generate cross-modal samples for expanding the training set. While these methods help address data sparsity, the quality of the generated samples is limited by the prior distribution of the pre-trained models, and their ability to augment extremely rare categories is constrained, often introducing distribution shift noise. The fourth approach involves the optimization of contrastive learning, as demonstrated by models such as FSCL \cite{yang2023neural}, which improves the contribution of sparse positive samples through weighted contrastive losses. However, the single-peak distribution assumption for feature spaces remains inconsistent with the multi-peak distribution relationships between modalities, complicating the modeling of complex semantic structures.

Overall, current methods face two key challenges in few-shot cross-modal retrieval tasks. First, the traditional single-peak distribution assumption struggles to capture the inherent multi-peak complexities of samples, leading to intra-modal bias, where sparse samples fail to adequately cover the multi-peak semantic distribution of categories. Second, conventional contrastive learning approaches are prone to the impact of distribution divergence under limited samples, resulting in inter-modal alignment bias, where the probability distribution differences between image and text feature spaces are amplified, undermining cross-modal semantic consistency. These issues collectively highlight the limitations of existing methods in multi-peak distribution modeling and cross-modal distribution constraints.

To overcome these challenges, we propose GCRDP, a model that leverages \textbf{G}aussian Mixture Models (GMM) and multi-positive sample \textbf{C}ontrastive learning to enhance comprehensive feature extraction. By enforcing \textbf{R}elative \textbf{D}istances \textbf{P}reservation between feature distributions of images and their corresponding text features, GCRDP improves the accuracy of cross-modal representations. We validate our approach through extensive experiments on four benchmark datasets in both zero-shot and few-shot settings, outperforming six state-of-the-art methods. The primary contributions of this paper include: 
\begin{itemize}
\item \textbf{Comprehensive Feature Extraction with GMM and Contrastive Learning (GC):} We propose a feature extraction framework that utilizes Gaussian Mixture Models to capture the multi-peak complex feature distributions of samples. By incorporating multi-positive contrastive learning, this framework provides deep semantic representations for few-shot learning, thereby improving the model's generalization ability.
\item \textbf{Cross-modal Relative Distance Preservation (RDP):} We introduce a cross-modal relative distance preservation constraint that aligns the relative distances between Gaussian components in the image feature space with those in the corresponding text feature space. This mechanism effectively narrows the semantic gap across modalities by enforcing distributional coherence between heterogeneous representations.
\item \textbf{Validation of Effectiveness:} We thoroughly evaluated our model on multiple few-shot (1,3,5) and zero-shot cross-modal retrieval tasks with four widely used datasets, including Wikipedia \cite{rasiwasia2010new}, Pascal Sentence \cite{rashtchian2010collecting}, NUS-WIDE \cite{wang2009nus}, and NUS-WIDE-10k \cite{wang2009nus}, outperforming six state-of-the-art methods.
\end{itemize}
\begin{figure*}[h]
    \centering
    \includegraphics[width=0.97\textwidth,height=7.9cm]{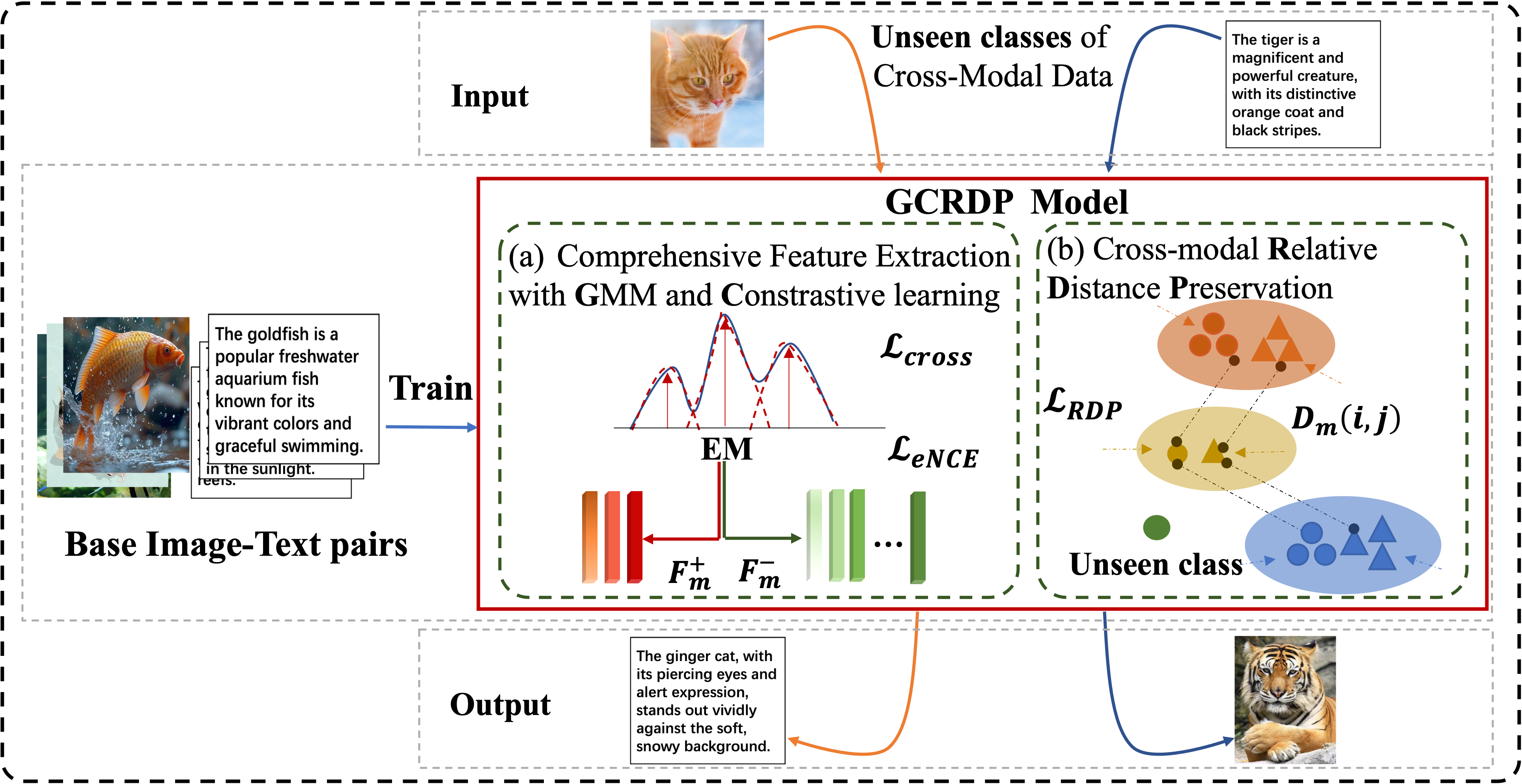}
    \caption{Schematic illustration of our proposed GCRDP. It consists of two parts: (a) Comprehensive Feature Extraction with \textbf{G}MM and multi-positive sample \textbf{C}ontrastive learning; (b) Cross-modal \textbf{R}elative \textbf{D}istance \textbf{P}reservation for few-shot learning. During the training phase, the model mitigates intra-modal bias and inter-modal bias by jointly minimizing the loss functions \begin{math}
        \mathcal{L}_{eNCE} + \alpha \mathcal{L}_{cross} + \lambda \mathcal{L}_{RDP}
    \end{math}. In the testing phase, the model demonstrates the ability to accurately retrieve the corresponding text or image for inputs from previously unseen classes (image or text).}
    \label{fig:gcrdp}
\end{figure*}
\section{Related Works}
\subsection{Few-shot Cross-modal Retrieval}
Few-shot cross-modal retrieval aims to address the challenge of unseen classes in training cross-modal models by integrating few-shot learning with cross-modal retrieval techniques. This approach learns semantic relationships between modalities with limited annotated samples, overcoming data scarcity issues \cite{wang2025cross}. Traditional supervised methods rely on large volumes of high-quality annotated data, limiting their performance in few-shot scenarios \cite{zhen2019deep,wang2024semantics,he2022category,zhang2023weighted,xie2024flexclipfeaturelevelgenerationnetwork}. Unsupervised methods, which do not require labeled data, have potential but generally do not match supervised learning in retrieval accuracy \cite{wang2024estimating,xiong2024reference,wang2022coder,pham2024composing}. Pre-training methods, which learn general modality representations and semantic alignment from large-scale unlabeled data before fine-tuning, have shown significant progress, especially with Transformer-based architectures \cite{lu2022cots,wang2023agree,zhang2024user,fu2024linguistic}. However, effective extraction of features from small-labeled datasets and improved retrieval accuracy in few-shot scenarios remain unsolved.
Recent advancements have introduced promising solutions. For example, Self-Others Net \cite{wang2021know} leverages semantic information from limited new class samples to enhance their features and improve retrieval performance. Similarly, Aligned Cross-Modal Memory \cite{huang2021few} aligns weakly supervised few-shot regions with corresponding words and updates cross-modal prototypes for more accurate similarity measurements.
\subsection{Contrastive Learning}
Contrastive learning is a key self-supervised learning paradigm that aligns semantically similar samples and separates dissimilar ones in embedding spaces using positive-negative pairs. Its development is characterized by three main aspects: data augmentation strategies, sample pair construction, and loss function design. Early works like MoCo \cite{9157636} introduced momentum encoders and dynamic memory banks, while SimCLR \cite{10.5555/3524938.3525087} emphasized large-batch training and composite data augmentation for InfoNCE loss. Later methods like SwAV \cite{10.5555/3495724.3496555} and BYOL \cite{Grill2020BootstrapYO} reduced the reliance on negative samples by using cluster prototypes or momentum encoders.
Contrastive learning has also achieved breakthroughs in multimodal and graph learning. CLIP \cite{radford2021learning} created a shared semantic space for zero-shot retrieval through contrast between image and text, and ALBEF \cite{Li2021AlignBeforeFuse} improved cross-modal interaction. In graph learning, Deep Graph Infomax \cite{48921} captured graph semantics by maximizing mutual information, and GraphMAE \cite{10.1145/3534678.3539321} used masked autoencoders for reconstruction.
This paper extends the InfoNCE loss to capture semantic relationships within the same image while distinguishing between different images by introducing multi-positive sample contrast.
\subsection{Prototype Construction}
Prototype construction aims to represent datasets using minimal representative prototypes \cite{snell2017prototypical}. In few-shot learning, prototype-based methods enhance model adaptability to novel classes \cite{chopin2024performance, he2023prototypeformer}. For example, in image classification, models adapt quickly to new classes by extracting key images as prototypes.
Global Average Pooling (GAP) \cite{yang2020prototype} is a common prototype construction technique that reduces feature maps to a global vector, but it may discard important spatial details, especially in high-resolution images.
Alternatively, Gaussian Mixture Models (GMM) \cite{zhao2023learning} model data as a mixture of Gaussian distributions, each representing a prototype. Using the Expectation-Maximization (EM) algorithm to estimate parameters, GMM captures complex data distributions more effectively. We employ GMM for prototype construction to achieve accurate feature representation.
\section{Method}
\subsection{Overview}
The primary challenge in few-shot cross-modal retrieval tasks is to effectively retrieve images or texts from unseen classes with only a limited number of labeled samples. To tackle this issue, we introduce a model named GCRDP, as depicted in Figure \ref{fig:gcrdp}. During the training phase, GCRDP employs a two-stage optimization strategy. In the first stage, a Gaussian Mixture Model (GMM) is utilized to model the joint probability density of image-text data, explicitly decoupling the multi-peak distribution characteristics of the samples. The modal component parameters are iteratively optimized using the Expectation-Maximization (EM) algorithm. Within this process, latent multi-peak clustering centers derived from the GMM serve as contrastive anchors to construct a multi-positive-sample contrastive learning task. Specifically, intra-class compactness constraints are imposed on different distribution modalities within the same semantic category, thereby enhancing the robustness of feature representations within individual modalities. In the second stage, a differentiable optimization objective is formulated by computing the relative distance between image and text feature distributions. Concretely, the relative distribution distance between image features and text features is constrained to maintain monotonic consistency with the cross-sample semantic similarity matrix. This approach transforms semantic alignment into a manifold matching problem in the feature distribution space. The mechanism adaptively adjusts the alignment strength for different few-shot categories, effectively mitigating divergence in probability densities across modalities. With these two stages, the model can classify, recognize, and retrieve inputs from previously unseen classes. We will provide a detailed introduction in the following sections.
\subsection{Comprehensive Feature Extraction with Gaussian Mixture Models and Contrastive Learning}
\begin{figure}[htbp]
    \centering
    \includegraphics[width=0.46\textwidth]{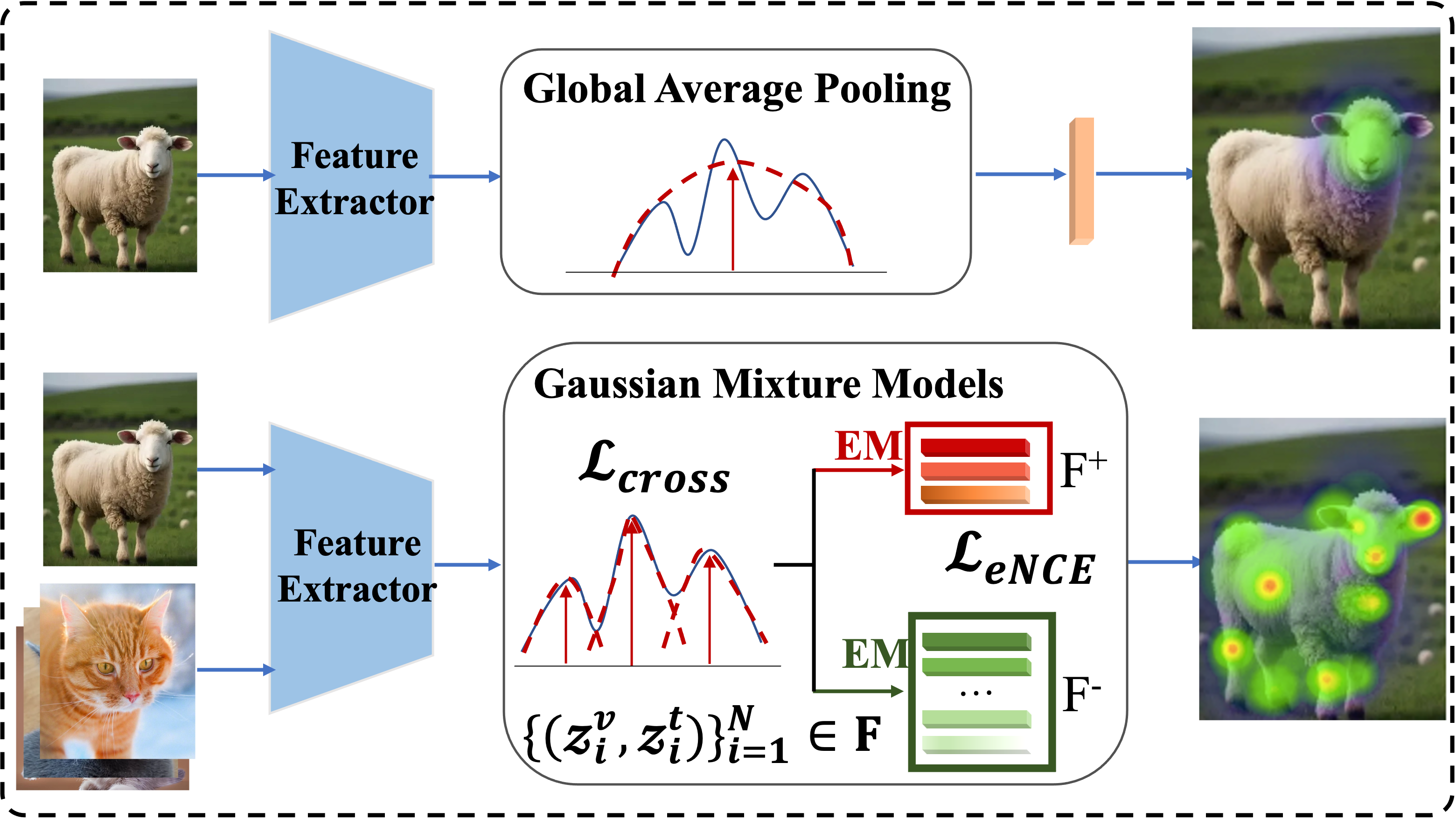}
    \caption{Comparison between the single prototype model (upper) based on Global Average Pooling and Gaussian Mixture Models (lower). The upper may cause the loss of some local detail information, while the lower can correlate diverse image regions, such as object parts, with multiple prototypes.}
    \label{fig:figures_GMM}
\end{figure}

Despite significant advancements in prototype-based models like Global Average Pooling (GAP) \cite{yang2020prototype} (as shown in the upper part of Figure \ref{fig:figures_GMM}), these methods still face challenges, particularly in overlooking certain feature distributions. Recent approaches have attempted to address this through prototype alignment \cite{9010855}, feature enhancement \cite{Nguyen2019FeatureWA}, and iterative mask refinement \cite{8954183}. However, the issue of semantic loss due to global average pooling persists.
In this paper, we introduce a comprehensive feature extraction method utilizing Gaussian Mixture Models (GMM). We initially extract features from both image and text modalities separately, projecting them into a shared latent space where GMM models their distribution. Each Gaussian component within the GMM represents a potential semantic unit adept at capturing the local semantics of both modalities. The strength of GMM lies in its capacity to model complex, multi-peak data distributions, facilitating the integration of local and global features across modalities.

Moreover, we propose an innovative multi-positive-sample contrastive learning mechanism based on Gaussian components. Given a batch of N image-text pairs\begin{math}
\{({z}_i^v,{z}_i^t)\}_{i=1}^N \in F
\end{math}, focusing on images as an example, each image feature \begin{math} {z}_i^v
\end{math} comprises \begin{math} K_v
\end{math} Gaussian components. One component is randomly selected as an anchor \begin{math}
z_{i,a}^v
\end{math} where \begin{math}
a\in \{{1,\ldots,K_v}\}
\end{math}, and the others within the same image from the positive sample set \begin{math}
 F^+
\end{math}. Negative samples \begin{math}
    N_{neg}
\end{math} are selected from Gaussian components of other \begin{math}
    N-1
\end{math} images \begin{math}
    \left\{{z}_{j_n}^v\right\}_{j_n\neq i} \in F^-
\end{math} (as shown in the lower part of Figure \ref{fig:figures_GMM}). By extending the InfoNCE loss for multi-positive-sample contrastive learning, this approach enhances intra-modal structure and component consistency while distinguishing semantic differences across instances.

\textbf{GMM model.} We define GMM as a probability mixture model that combines base distribution probabilities linearly as follows:
\begin{equation}
p(f_m|\theta_m)=\sum_{k=1}^{K_m}\pi_{m,k}\mathcal{N}(f_m|\mu_{m,k},\sum_{m,k})
\end{equation}
where \begin{math}
    m \in \{{v,t}\}
\end{math} denotes the modality of image or text, \begin{math}
    \pi_{m,k}
\end{math} are the mixing weights with \begin{math}
    0\le\pi_{m,k}\le1
\end{math} and \begin{math}
\sum_{k=1}^{K}\pi_{m,k}=1
\end{math}. The model parameters, \begin{math}
\theta_m = \{{\pi_{m,k}, \mu_{m,k}, \sum_{m,k}}\}_{k=1}^K
\end{math}, are learned during GMM estimation, where \begin{math}
    \mathcal{N}(f_m|\mu_{m,k}, \sum_{m,k})
\end{math} represents the probability density function of k-th
Gaussian component, with mean \begin{math}
    \mu_{m,k}
\end{math} and covariance matrix \begin{math}
    \sum_{m,k}
\end{math}. 

\textbf{Model learning.} During training, we estimate GMM using the Expectation-Maximization (EM) algorithm, as depicted in the lower part of Figure \ref{fig:figures_GMM}. This process iteratively involves E-steps and M-steps. In the E-step, given current model parameters, we compute the posterior probability for each sample \(f_{m,n}\):
\begin{equation}
\gamma_{m,k}^{(t)}(f_{m,n})\ =\ \frac{\pi_{m,k}^{(t)}\mathcal{N}(f_{m,n}|\mu_{m,k}^{(t)},\sum_{m,k}^{(t)})}{\sum_{j=1}^{K_m}\pi_{m,j}^{(t)}\mathcal{N}(f_{m,n}|\mu_{m,j}^{(t)},\sum_{m,j}^{(t)})}
\end{equation} where \begin{math}
    \gamma_{m,k}^{(t)}(f_{m,n})
\end{math} is the probability that \(f_{m,n}\) belongs to the k-th Gaussian component at iteration t. 

The M-step updates the parameters as follows:
\begin{equation}
\pi_{m,k}^{(t+1)}\ =\ \frac{1}{N}\sum_{n=1}^{N}{\gamma_{m,k}^{(t)}(f_{m,n})}
\end{equation} where it is used to update the mixing weight of the k-th Gaussian component.

\begin{equation}
\mu_{m,k}^{(t+1)}\ =\ \frac{\sum_{n=1}^{N}{\gamma_{m,k}^{(t)}(f_{m,n})f_{m,n}}}{\sum_{n=1}^{N}{\gamma_{m,k}^{(t)}(f_{m,n})}}
\end{equation} where it is used to update the mean of the k-th Gaussian component.

\begin{equation}
\sum_{m,k}^{(t+1)}\ =\ \frac{\sum_{n=1}^{N}{\gamma_{m,k}^{(t)}(f_{m,n})}(f_{m,n}\ -\ \mu_{m,k}^{(t+1)})(f_{m,n}\ -\ \mu_{m,k}^{(t+1)})^\intercal}{\sum_{n=1}^{N}{\gamma_{m,k}^{(t)}(f_{m,n})}}
\end{equation} where it is used to update the covariance matrix of the k-th Gaussian component.

For each sample, we select the Gaussian component with the highest posterior probability:
\begin{equation}
{z}_m = \text{L2-Norm}\left(\left[\mu_{m,k^*,\ } \sqrt{\text{diag}(\Sigma_{m,k^*})}, \pi_{m,k^*,\ }\right]\right)
\end{equation}

The cross-modal joint feature is the concatenation of image and text features:
\begin{equation}
{z}\ =\ {z}_v\oplus{z}_t
\end{equation}

To capture intra-image semantic relationships while distinguishing inter-image variations, we extend the InfoNCE loss:\begin{equation}
\begin{split}
\mathcal{L}_{\text{eNCE}} &= -\frac{1}{K_v - 1} \\
&\sum_{k \neq a} \log \left( \frac{\exp\left(s(z_{i,a}^v, z_{i,k}^v) / \tau\right)}{\exp\left(s(z_{i,a}^v, z_{i,k}^v) / \tau\right) + \sum_{n=1}^{N_{\text{neg}}} \exp\left(s(z_{i,a}^v, z_{j_n}^v) / \tau\right)} \right)
\end{split}
\end{equation} where \begin{math}
    s\left(\cdot,\cdot\right)
\end{math} is the cosine similarity function, and \begin{math}
    \tau
\end{math} is a temperature coefficient that controls the scaling of similarity scores. The same approach applies to the text modality. The total loss is the sum of both modalities' contrastive losses. To ensure semantic consistency between modalities, we add a cross-modal similarity regularization term: 
\begin{equation}
\mathcal{L}_{\text{cross}} = \sum_{i=1}^{N} \sum_{k=1}^{K_v} \sum_{l=1}^{K_t} \left\| {z}_{i,k}^v, {z}_{i,l}^t - XA(k = l) \right\|_2^2 
\end{equation}
where XA(k=\(l\)) is an indicator function aligning the k-th Gaussian component of an image with the \(l\)-th component of the corresponding text, as shown in Figure \ref{fig:figures_GMM}.
\subsection{Cross-Modal Relative Distance Preservation}
\begin{figure}[htbp]
    \centering
    \includegraphics[width=0.46\textwidth]{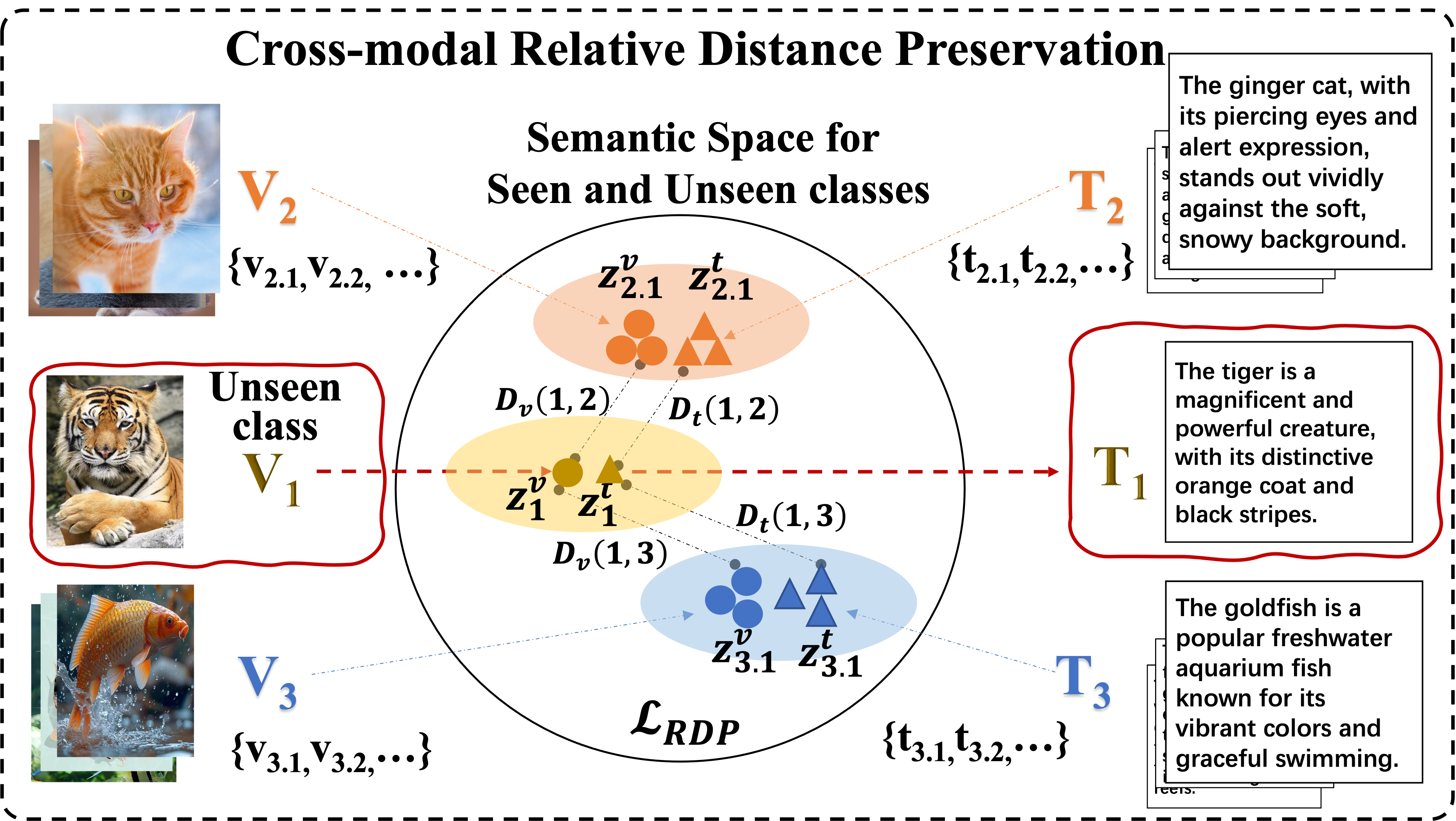}
    \caption{Cross-modal relative distance preservation. By enforcing the consistency of the feature distribution among image Gaussian components \begin{math}
        {z}_i^v
    \end{math} with that among the corresponding text Gaussian components \begin{math}
        {z}_i^t
    \end{math}, cross-modal semantics are aligned at a fine-grained level, where \begin{math}
        D_v
    \end{math} denotes the distance between image Gaussian components, and \begin{math}
        D_t
    \end{math} denotes the distance between text Gaussian components. The \begin{math}
        \mathcal{L}_{RDP}
    \end{math} is the loss function.}
    \label{fig:figures_RDP}
\end{figure}
In scenarios where training data is limited, instance-level contrastive learning often overfits the available sample pairs, impairing generalization to unseen local semantic relationships. To solve this problem, we introduce the Cross-modal Relative Distance Preservation (RDP) constraint, as depicted in Figure \ref{fig:figures_RDP}. This constraint aims to align cross-modal semantics at a fine-grained level by ensuring consistency between the feature distributions of image Gaussian components and their corresponding text Gaussian components. This alignment enhances the model's robustness against data scarcity and modality disparities. Specifically, for each instance, the data across different modalities should not only share similarities in global features but also maintain the alignment of their internal semantic components within the embedding space. However, in few-shot settings, the challenge of learning such fine-grained cross-modal alignments with limited samples can lead to overfitting.

To implement the RDP constraint, we first calculate intra-modal similarity matrices for both image and text modalities within each training batch, which consists of a set of image-text pairs \begin{math}
    {\{({z}}_i^v,{z}_i^t)\}_{i=1}^N
\end{math}. The similarity matrices are defined as:
\begin{equation}
    D_m\in\mathbb{R}^{N\times N},D_m(i,j)\ =\ s({z}_i^m,{z}_j^m)
\end{equation} where \begin{math}
    m \in \{v,t\}
\end{math} denotes the image or text modalities, and \begin{math}
    s(\cdot,\cdot)
\end{math} is the similarity metric between feature vectors.

We then enforce the alignment of similarity distributions between image pairs and their corresponding text pairs using the RDP loss function \begin{math}
    \mathcal{L}_{RDP}
\end{math}, as shown in Figure \ref{fig:figures_RDP}. To reduce the impact of noisy samples, this constraint is applied only to high-confidence pairs: \begin{equation}
\mathcal{L}_{\text{RDP}} = \frac{\sum_{i,j} XA(D_v(i,j) > \theta) \cdot \left\| D_v(i,j) - D_t(i,j) \right\|_2^2}{\sum_{i,j} XA(D_v(i,j) > \theta)} 
\end{equation}
where \begin{math}
    \theta=0.5
\end{math} is the similarity threshold. This loss ensures that if image samples i and j are similar in the image feature space, their corresponding text samples should also show similar traits in the text feature space. Conversely, dissimilar image samples should lead to reduced similarity in the corresponding text samples.

The overall optimization of the model is achieved by jointly minimizing the following loss function:
\begin{equation}
    \mathcal{L}_{total}\ =\ \mathcal{L}_{eNCE}\ +\alpha\mathcal{L}_{cross}+\lambda\mathcal{L}_{RDP}
\end{equation}
where \begin{math}
\alpha
\end{math}
and \begin{math}
\lambda
\end{math} are hyperparameters that balance the influence of each loss term, as shown in Figure \ref{fig:gcrdp}. By optimizing these losses concurrently, the model not only secures global matching across modalities but also ensures consistency in their local structures. This approach markedly boosts retrieval accuracy and generalization in few-shot cross-modal retrieval tasks.

\section{Experiment}
\subsection{Datasets}
We conducted experiments on four widely used benchmark datasets: Wikipedia \cite{rasiwasia2010new}, Pascal Sentence \cite{rashtchian2010collecting}, NUS-WIDE \cite{wang2009nus}, and NUS-WIDE-10k \cite{wang2009nus}. These datasets were partitioned into source and target domains, where the source domain encompasses training data from multiple classes, and the target domain includes test data from classes not present in the source domain. The target classes were either unseen or minimally represented during model training. For our evaluation, we employed the same data partitioning as in Flexclip \cite{xie2024flexclipfeaturelevelgenerationnetwork}. The specifics of each dataset are described below:

 \textbf{Wikipedia:} Comprises 2,866 image-text pairs sourced from Wikipedia, each belonging to one of ten classes. The source domain includes 2,173 pairs from five randomly chosen classes, while the target domain consists of 693 pairs from five distinct classes.
 \textbf{NUS-WIDE-10k:} A reduced version of NUS-WIDE with 10,000 image-text pairs. We randomly picked 1,000 pairs from the ten most frequent classes of NUS-WIDE for this dataset. The source domain contains 8,000 pairs from five classes, with the target domain having 2,000 pairs from five different classes.
\textbf{Pascal Sentence:} Contains 1,000 images, each assigned to one of twenty classes. The source domain features 800 pairs from ten randomly selected classes, whereas the target domain comprises 200 pairs from ten other classes.
 \textbf{NUS-WIDE:} Encompasses 270,000 images with tags, each linked to at least one of 81 classes. We selected 42,859 pairs from the ten most common classes to form our subset. Here, the source domain includes 25,685 pairs from five classes, and the target domain has 12,174 pairs from five different classes.

\subsection{Metrics}
Following previous cross-modal retrieval methods \cite{xie2024flexclipfeaturelevelgenerationnetwork}, we evaluate the performance of image-to-text and text-to-image retrieval tasks using the mean average precision (mAP) score. The average precision (AP) for each query is computed as follows:
\begin{equation}
    AP\ =\ \frac{1}{T}\sum_{r=1}^{R}{P_r\times\delta\left(r\right)}
\end{equation}
where R is the set of retrieved items, T is the number of relevant items among the retrieved items, \begin{math}
    P_r
\end{math} denotes the precision of the top r retrieved items, and \begin{math}
    \delta_r
\end{math} is an indicator function. We use cosine distance to measure the similarity between feature vectors. Specifically, if the r-th item is relevant to the query, the value of \begin{math}
    \delta_r
\end{math}  is set to 1. The mean average precision (mAP) is then calculated as the mean of the average precision scores across all queries:
\begin{equation}
    mAP = \frac1{n}\sum_{i=1}^{n}{AP(i)}
\end{equation} where n is the number of the queries.

\begin{table*}[h]
    \centering
    \caption{Comparisons of our model and six compared methods on four benchmark datasets for zero-shot, 1-shot, 3-shot, and 5-shot cross-modal retrieval. The best results are marked in bold, the second-best results are marked in blue, and the third-best results are marked in green, respectively. The 4 experiment results are the same in CLIP, so the k-shot information is "-".}
    \label{tab:results}
    \begin{tabular}{lccccccccccccc}
        \toprule
        \multirow{2}{*}{Model} & \multirow{2}{*}{k-shot} & \multicolumn{3}{c}{Wikipedia} & \multicolumn{3}{c}{NUS-WIDE-10k} & \multicolumn{3}{c}{Pascal Sentence}  & \multicolumn{3}{c}{NUS-WIDE} \\
         \cmidrule(lr){3-5} \cmidrule(lr){6-8} \cmidrule(lr){9-11} \cmidrule(lr){12-14}
         & & I2T  & T2I  & Avg & I2T  & T2I  & Avg& I2T  & T2I  & Avg& I2T  & T2I  & Avg\\
        \midrule  
        \multirow{4}{*}{MRL \cite{hu2021learning} } 
        & zero-shot & 41.8 & 45.5 & 43.7 & 45.4 & 45.2 & 45.3 & 51.9 & 50.3 & 51.1 & 49.8 & 49.1 & 49.4 \\
        & 1-shot & 42.7 & 46.1 & 44.4 & 45.9 & 44.9 & 45.4 & 52.9 & 51.7 & 52.3 & 49.2 & 49.1 & 49.1 \\
        & 3-shot & 43.3 & 47.5 & 45.4 & 47.0 & 45.7 & 46.4 & 56.6 & 61.0 & 58.8 & 50.6 & 50.3 & 50.4 \\
        & 5-shot & 41.2 & 48.0 & 44.6 & 47.1 & 45.7 & 46.4 & 61.9 & 61.0 & 61.5 & 50.9 & 48.9 & 49.9 \\
        \midrule
        \multirow{4}{*}{CFSA \cite{xu2020correlated}} 
        & zero-shot & 42.1 & 35.6 & 38.8 & 38.3 & 40.2 & 39.2 & 46.3 & 44.0 & 45.1 & 45.0 & 47.3 & 46.1 \\
        & 1-shot  & 40.0 & 34.8 & 37.4 & 38.1 & 40.3 & 39.2 & 49.5 & 49.1 & 49.3 & 45.1 & 47.1 & 46.1 \\
        & 3-shot & 40.9 & 34.9 & 37.9 & 37.1 & 39.3 & 38.2 & 57.7 & 57.8 & 57.8 & 45.0 & 47.0 & 46.0 \\
        & 5-shot & 40.4 & 35.2 & 37.8 & 37.8 & 39.8 & 38.8 & 63.6 & 65.4 & 64.5 & 45.8 & 47.7 & 46.7 \\
        \midrule
        \multirow{4}{*}{JFSE \cite{xu2022joint} } 
        & zero-shot & 42.0 & 35.2 & 38.6 & 37.9 & 40.3 & 39.1 & 43.8 & 42.3 & 43.1 & 45.8 & 47.3 & 46.5 \\
        & 1-shot & 40.5 & 34.9 & 37.7 & 38.0 & 40.7 & 39.4 & 52.3 & 51.7 & 52.0 & 44.7 & 46.7 & 45.7 \\
        & 3-shot & 39.3 & 33.7 & 36.5 & 38.2 & 39.9 & 39.1 & 57.8 & 57.7 & 57.7 & 45.0 & 46.9 & 46.0 \\
        & 5-shot & 41.4 & 35.0 & 38.2 & 37.8 & 39.9 & 38.8 & 62.2 & 64.7 & 63.5 & 44.7 & 46.4 & 45.6 \\
        \midrule
        \multirow{4}{*}{MDVAE \cite{tian2022multimodal}} 
        & zero-shot & 47.4 & 45.9 & 46.6 & 40.0 & 42.5 & 41.2 & 47.0 & 46.7 & 46.9 & 39.9 & 42.4 & 41.1 \\
        & 1-shot & 45.0 & 43.3 & 44.2 & 41.8 & 43.7 & 42.8 & 44.8 & 45.5 & 45.1 & 39.7 & 41.9 & 40.8 \\
        & 3-shot & 47.6 & 44.5 & 46.0 & 40.3 & 43.4 & 41.8 & 49.1 & 48.9 & 49.0 & 39.6 & 41.0 & 40.3 \\
        & 5-shot & 49.7 & 45.0 & 47.3 & 41.0 & 44.4 & 42.7 & 52.5 & 50.8 & 51.6 & 40.7 & 41.8 & 41.3 \\
        \midrule
        CLIP \cite{radford2021learning} &   - &\textcolor{green}{52.6} &\textcolor{green}{47.9} &\textcolor{green}{50.2} &\textcolor{green}{48.0} &\textcolor{green}{52.6} &\textcolor{green}{50.3} &\textcolor{green}{66.9} &\textcolor{green}{65.9} &\textcolor{green}{66.4} &\textcolor{green}{58.1} &\textcolor{green}{60.2} &\textcolor{green}{59.2} \\
        \midrule
        \multirow{4}{*}{FLEX-CLIP \cite{xie2024flexclipfeaturelevelgenerationnetwork}} 
        & zero-shot &\textcolor{blue}{53.9}  &\textcolor{blue}{51.2} &\textcolor{blue}{52.6} &\textcolor{blue}{56.9} &\textcolor{blue}{58.9} &\textcolor{blue}{57.9} &\textcolor{blue}{69.0} &\textcolor{blue}{68.9} &\textcolor{blue}{68.9} &\textcolor{blue}{66.1} &\textcolor{blue}{68.0} &\textcolor{blue}{67.1} \\
        & 1-shot &\textcolor{blue}{55.1} &\textcolor{blue}{52.3} &\textcolor{blue}{53.7} &\textcolor{blue}{57.3} &\textcolor{blue}{59.2} &\textcolor{blue}{58.2} &\textcolor{blue}{69.5} &\textcolor{blue}{69.0} &\textcolor{blue}{69.3} &\textcolor{blue}{66.0} &\textcolor{blue}{67.7} &\textcolor{blue}{66.8} \\
        & 3-shot &\textcolor{blue}{59.4} &\textcolor{blue}{55.4} &\textcolor{blue}{57.4} &\textcolor{blue}{60.9} &\textcolor{blue}{62.6} &\textcolor{blue}{61.8} &\textcolor{blue}{71.3} &\textcolor{blue}{71.8} &\textcolor{blue}{71.5} &\textcolor{blue}{66.6} &\textcolor{blue}{68.7} &\textcolor{blue}{67.6} \\
        & 5-shot &\textcolor{blue}{56.5} &\textcolor{blue}{53.3} &\textcolor{blue}{54.9} &\textcolor{blue}{58.5} &\textcolor{blue}{60.3} &\textcolor{blue}{59.4} &\textcolor{blue}{70.1} &\textcolor{blue}{70.0} &\textcolor{blue}{70.0} &\textcolor{blue}{66.2} &\textcolor{blue}{68.0} &\textcolor{blue}{67.1} \\
        \midrule
        \multirow{4}{*}{GCRDP (Ours)}
        & zero-shot & \textbf{56.6} & \textbf{54.3} & \textbf{55.4} & \textbf{58.8} & \textbf{61.4} & \textbf{60.1} & \textbf{71.3} & \textbf{70.7} & \textbf{71.0} & \textbf{69.3} & \textbf{70.4} & \textbf{69.9} \\
        & 1-shot & \textbf{57.5} & \textbf{54.7}  & \textbf{56.0}  & \textbf{60.8}  & \textbf{62.1}  & \textbf{61.4}  & \textbf{72.3}  & \textbf{71.6}  & \textbf{71.9}  & \textbf{68.4}  & \textbf{70.2}  & \textbf{69.3}  \\
        & 3-shot & \textbf{62.2} & \textbf{58.9} & \textbf{60.5} & \textbf{63.5} & \textbf{64.9} & \textbf{64.2} & \textbf{74.1} & \textbf{73.5} & \textbf{73.8} & \textbf{68.9} & \textbf{70.6} & \textbf{69.7} \\
        & 5-shot & \textbf{59.3} & \textbf{55.8} & \textbf{57.5} & \textbf{61.2} & \textbf{63.4} & \textbf{62.3} & \textbf{72.9} & \textbf{72.3} & \textbf{72.6} & \textbf{68.5} & \textbf{70.8} & \textbf{69.6} \\
   \bottomrule
    \end{tabular}
\end{table*}
\subsection{Implementation Details}
In our model, the key parameters of the Gaussian Mixture Model (GMM) include the number of components, mixture weights, mean vectors, and covariance matrices. These parameters are optimized using the Expectation-Maximization (EM) algorithm. In our experiments, we typically set the number of components to 3 to capture the complex distribution of multimodal data. The mixture weights conform to the normalization condition, indicating the relative importance of each component within the mixture. The mean vectors and covariance matrices define the location and dispersion of each component, respectively. By adjusting these parameters, GMM effectively adapts to complex, multi-peak distributions, enhancing its capability to handle feature distributions in few-shot scenarios. During training, we utilize the Adam optimizer to achieve efficient convergence and stable optimization, with a learning rate set at $1 \times 10^{-4}$.

To assess our model, we conducted experiments on both image-to-text (I2T) and text-to-image (T2I) tasks under zero-shot and few-shot (1, 3, 5-shot) conditions. We benchmarked our proposed GCRDP method against six state-of-the-art approaches on retrieval tasks, including MRL \cite{hu2021learning}, CFSA \cite{xu2020correlated}, JFSE \cite{xu2022joint}, MDVAE \cite{tian2022multimodal}, CLIP \cite{radford2021learning}, and FLEX-CLIP \cite{xie2024flexclipfeaturelevelgenerationnetwork}. 

\subsection{Comparisons Results}
As shown in Table 1, our proposed method GCRDP outperforms six state-of-the-art methods comprehensively in both zero-shot and few-shot (1/3/5-shot) scenarios. Across the four major benchmark datasets, this method demonstrates significant performance superiority.

\subsection{Ablation Studies}
\textbf{Benchmark Comparison Analysis:} Baseline methods can be categorized into two classes: MRL is the conventional method, whereas CFSA, JFSE, MDVAE, CLIP, and FLEX-CLIP are recent few-shot optimization methods. Emerging methods significantly outperform traditional baselines through tailored designs. Among these, CLIP and its extension framework, FLEX-CLIP, stand out prominently. CLIP achieves feature enhancement and cross-modal alignment through large-scale pre-training, while FLEX-CLIP effectively addresses issues like data imbalance, feature degradation, and modal differences in few-shot cross-modal retrieval by employing a VAE-GAN network and gated residual networks.
\begin{table*}[htbp]
    \centering
    \caption{The effectiveness of the ablation experiment on the GCRDP model.}
    \label{tab:results}
    \begin{tabular}{lccccccccccccccc}
        \toprule
 \multirow{2}{*}{Model} & \multirow{2}{*}{k-shot} & \multicolumn{3}{c}{Wikipedia} & \multicolumn{3}{c}{NUS-WIDE-10k} & \multicolumn{3}{c}{Pascal Sentence}  & \multicolumn{3}{c}{NUS-WIDE} \\
         \cmidrule(lr){3-5} \cmidrule(lr){6-8} \cmidrule(lr){9-11} \cmidrule(lr){12-14}
         & & I2T  & T2I  & Avg & I2T  & T2I  & Avg& I2T  & T2I  & Avg& I2T  & T2I  & Avg\\
        \midrule  
        \multirow{3}{*}{Full}
        &zero       & 56.6 & 54.3 & 55.4 & 58.8 & 61.4 & 60.1 & 71.3 & 70.7 & 71.0 & 69.3 & 70.4 & 69.9 \\
        & 1-shot & 57.5 & 54.7 & 56.0 & 60.8 & 62.1 & 61.4 & 72.3 & 71.6 & 71.9 & 68.4 & 70.2 & 69.3 \\
        & 3-shot & 62.2 & 58.9 & 60.5 & 63.5 & 64.9 & 64.2 & 74.1 & 73.5 & 73.8 & 68.9 & 70.6 & 69.7 \\
        & 5-shot & 59.3 & 55.8 & 57.5 & 61.2 & 63.4 & 62.3 & 72.9 & 72.3 & 72.6 & 68.5 & 70.8 & 69.6 \\
        \midrule
        \multirow{3}{*}{without GMM } 
        &zero & 54.2 & 52.8 & 53.5 & 56.9 & 59.1 & 58.0 & 69.9 & 68.2 & 69.0 & 67.6 & 68.1 & 67.8 \\
        & 1-shot & 55.7 & 53.3 & 54.5 & 58.4 & 59.8 & 59.1 & 70.6 & 69.8 & 70.2 & 66.1 & 68.0 & 67.0 \\
        & 3-shot & 59.5 & 56.4 & 57.9 & 61.2 & 62.6 & 61.9 & 71.4 & 70.6 & 71.0 & 67.2 & 69.1 & 68.1 \\
        & 5-shot & 57.9 & 54.1 & 55.9 & 59.3 & 61.7 & 60.5 & 71.2 & 70.1 & 70.6 & 66.8 & 68.9 & 67.8 \\
        \midrule
        \multirow{3}{*}{without RDP} 
        &zero & 54.6 & 52.5 & 53.5 & 56.3 & 58.7 & 57.4 & 69.8 & 67.7 & 68.7 & 67.1 & 67.9 & 67.5 \\ 
        & 1-shot & 55.5 & 52.9 & 54.1 & 57.8 & 59.6 & 58.7 & 70.2 & 70.1 & 70.1 & 65.6 & 68.2 & 66.9 \\
        & 3-shot & 59.9 & 57.2 & 58.5 & 60.7 & 62.9 & 61.8 & 71.0 & 71.0 & 71.0 & 66.8 & 69.1 & 67.9 \\ 
        & 5-shot & 56.7 & 52.8 & 54.7 & 59.7 & 61.2 & 60.4 & 70.2 & 70.5 & 70.3 & 66.0 & 67.9 & 66.9 \\
   \bottomrule
    \end{tabular}
\end{table*}

\textbf{Key Experimental Results Analysis:} 
\textit{1) Zero-shot Scenario:} Our method outperforms FLEX-CLIP, the current best baseline, by 2.8, 2.2, 2.1, and 2.8 percentage points on the Wikipedia, NUS-WIDE-10k, Pascal Sentence, and NUS-WIDE datasets, respectively. Notably, on the Pascal Sentence dataset, which benefits from balanced category distribution and straightforward semantics, our method achieves optimal cross-modal alignment with scores of I2T 71.3 and T2I 70.7.

\textit{2) 1-shot Scenario:} Our approach surpasses the average score of 60 on the NUS-WIDE-10k, Pascal Sentence, and NUS-WIDE datasets. Specifically, Pascal Sentence leads with I2T 72.3 and T2I 71.6, averaging at 71.9. On the Wikipedia dataset, where modal alignment noise increases due to text redundancy, our method still improves by 2.3 percentage points over FLEX-CLIP, averaging at 56.

\textit{3) 3-shot Scenario:} Our method outperforms CLIP by approximately 10 percentage points across all metrics (I2T/T2I/Avg) on all four datasets and leads FLEX-CLIP by an average of 3 percentage points. This configuration strikes a balance between sample sufficiency and the risk of overfitting, proven to be the optimal few-shot setup.

\textit{4) 5-shot Scenario:} Here, sample quality becomes a critical variable. On the Wikipedia dataset, complexity in text leads to a decrease in the 5-shot average score by 2.5 percentage points compared to 3-shot, while on the NUS-WIDE dataset, due to high intra-class similarity and strong inter-class distinction, performance across different shot configurations tends to converge.

\begin{figure*}[htbp]
    \centering
    \includegraphics[width=0.9\textwidth]{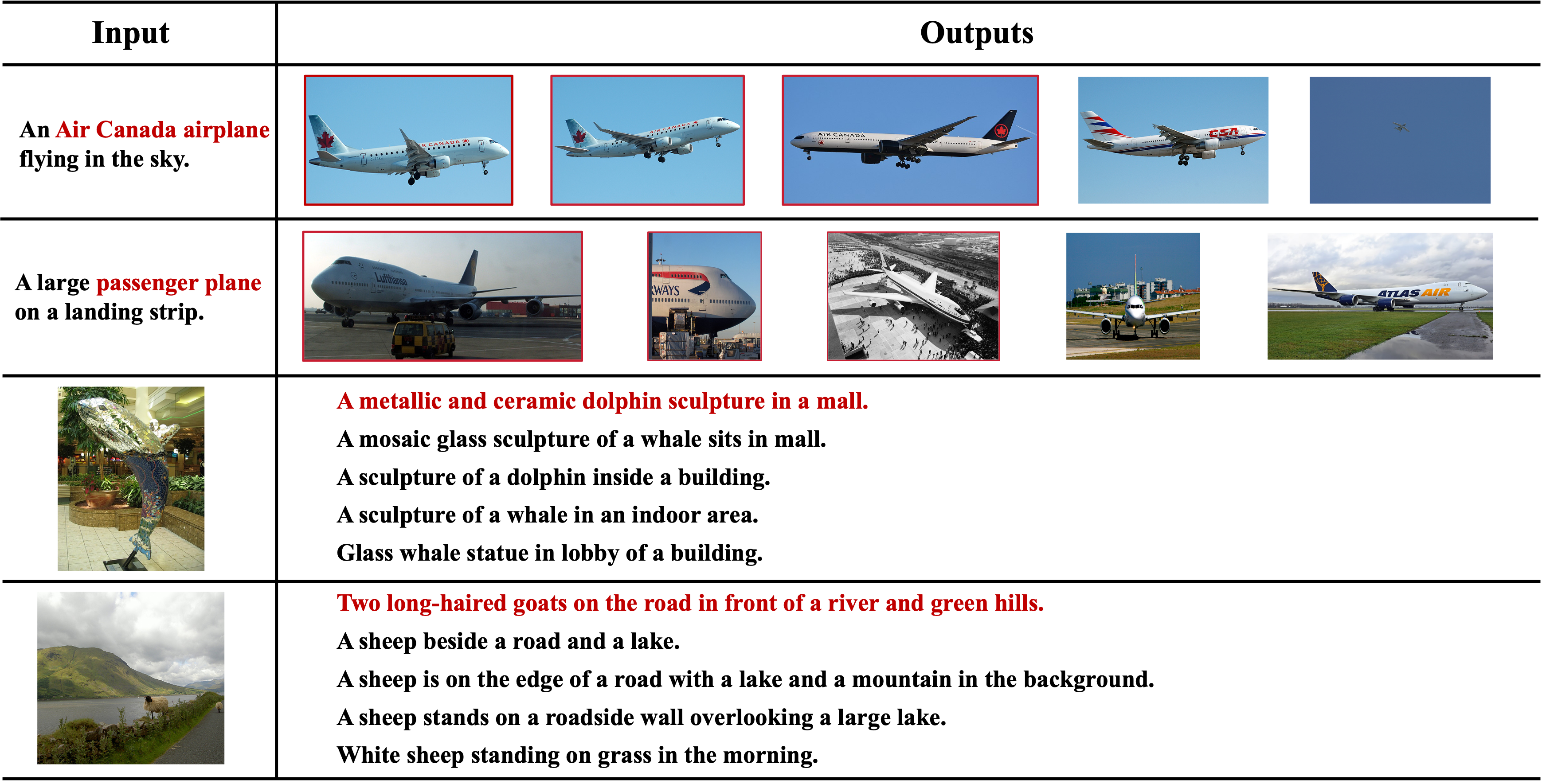}
    \caption{The examples of top-5 cross-modal retrieval results of our GCRDP.}
    \label{fig:visual}
\end{figure*}

\textit{5) Case Study:} In addition to the quantitative results, we also provide qualitative examples of few-shot cross-modal retrieval using our GCRDP model. As illustrated in Figure \ref{fig:visual}, the samples are drawn from the target domain in Wikipedia. For each input sentence, the top 5 retrieved images are displayed, and for each input image, the top 5 retrieved sentences are shown. The correctly retrieved candidates are marked in red. These examples demonstrate that the combination of comprehensive feature extraction and relative distance preservation is effective in improving retrieval performance.

Three key factors influence the performance of few-shot cross-modal retrieval: the dynamic balance of sample quantity and quality, the semantic complexity and noise levels of the dataset, and the model's robustness in feature extraction. Our method, through a comprehensive feature extraction and relative distance preservation mechanism, demonstrates more robust cross-modal alignment characteristics in complex scenarios.
To evaluate the effectiveness of the Gaussian Mixture Model (GMM) and Relative Distance Preservation (RDP) within the GCRDP model, we executed ablation studies across four datasets under zero-shot, 1-shot, 3-shot, and 5-shot settings by excluding each component individually. The performance comparison is presented in Table 2.

\textbf{GMM.} GMM models the latent distributions of cross-modal data. Its exclusion, as shown in Table 2, results in a performance decline across all datasets and conditions, affirming its significant contribution to model performance. In the Wikipedia dataset, the 3-shot average decreases from 60.5 to 57.9 (a 2.6 points reduction), and the 5-shot average from 57.5 to 55.9 (1.6 points), suggesting a more substantial impact in the 3-shot scenario. This could be attributed to GMM's ability to manage sparse alignments between modalities by modeling mixed distributions. For the NUS-WIDE-10k dataset, the 3-shot average drops from 64.2 to 61.9 (a 2.3 points decrease), and the 5-shot from 62.3 to 60.5 (1.8 points). In the NUS-WIDE dataset, zero-shot and 1-shot performance decreases are 2.1 and 2.3 points, respectively, with 3-shot and 5-shot reductions at 1.6 and 1.8 points. These observations indicate that GMM is particularly effective at handling complex multi-label data by enhancing modality alignment. In the Pascal Sentence dataset, the 3-shot average falls by 2.8 points from 73.8 to 71.0, marking the largest observed decrease, likely due to GMM's capacity to capture fine-grained subclass distributions. Collectively, these findings confirm that GMM improves the model's adaptability to complex cross-modal data, especially under low-shot conditions, potentially reducing the impact of low-quality samples like long-tail text descriptions in Wikipedia through probabilistic weighting.

\textbf{Relative Distance Preservation (RDP).} RDP concentrates on modeling the relational distribution between cross-modal samples to regulate the embedding space. The omission of RDP leads to performance reductions across all datasets and conditions, highlighting its role in enhancing model capabilities. In the Wikipedia dataset, the 5-shot average drops from 57.5 to 54.7 (a 2.8 points decrease), which is more pronounced than the 3-shot drop from 60.5 to 58.5 (2.0 points). For the NUS-WIDE dataset, the 5-shot average decreases from 69.6 to 66.9 (a 2.7 points decrease), while the 3-shot average falls from 69.7 to 67.9 (1.8 points), suggesting that RDP's regularization effect becomes more evident with larger sample sizes to prevent overfitting. In the Pascal Sentence dataset, the 1-shot T2I performance declines from 71.6 to 70.1 (a 1.5 points decrease), and the 3-shot T2I drops from 73.5 to 71.0 (2.5 points), indicating an enhanced benefit from RDP as sample numbers increase. Thus, RDP is vital for managing relationship distributions to mitigate overfitting, particularly with expanded sample sets.

\textbf{The Contribution and Role of GMM and RDP in the Complete Model.} Analysis of these results reveals that GMM significantly impacts model performance, with its absence leading to a more substantial performance drop compared to RDP. Nevertheless, both GMM and RDP yield complementary advantages. GMM addresses the intricacies of data distribution, managing issues such as multi-peak distribution scenarios and noise, while RDP tackles the complexity of relationships, including sparse sample similarities. For instance, in the Pascal Sentence dataset, both methods show considerable influence (-2.8 vs. -2.5), underlining the necessity of concurrently addressing data distribution and inter-sample relations.
\section{Conclusion}
In this paper, we propose GCRDP, a robust and efficient methodology for few-shot cross-modal retrieval. By employing a Gaussian Mixture Model (GMM), we comprehensively capture the complex distribution of samples, providing richer semantic information crucial for few-shot scenarios. Additionally, we introduce a novel semantic alignment strategy that ensures consistency between the similarity distances of text embeddings and corresponding image embeddings. This effectively bridges the semantic gap between different modalities, significantly enhancing the accuracy of cross-modal retrieval. Extensive experiments and analyses across multiple datasets demonstrate the superiority of our method.







\bibliography{GCRDP/GCRDP}

\end{document}